\documentclass[10pt,twocolumn,letterpaper]{article}

\usepackage{cvpr}
\usepackage{times}
\usepackage{epsfig}
\usepackage{graphicx}
\usepackage{amsmath}
\usepackage{amssymb}
\usepackage{tabulary}
\usepackage{multirow}
\usepackage{fancyhdr}

\usepackage{soul}

\usepackage[bookmarks=false,colorlinks=true,linkcolor=black,citecolor=black,filecolor=black,urlcolor=black]{hyperref}

\usepackage[british,UKenglish,USenglish,english,american]{babel}

\usepackage{amssymb}
\usepackage{amsthm}
\usepackage[table, dvipsnames]{xcolor}
\usepackage{subfigure}
\usepackage{amsmath}
\newtheorem{theorem}{Theorem}
\newtheorem{lemma}{Lemma}
\usepackage[ruled]{algorithm2e}
\usepackage{algorithm,algpseudocode,float}
\usepackage{lipsum}
\makeatletter
\newenvironment{breakablealgorithm}
  {% \begin{breakablealgorithm}
   \begin{center}
     \refstepcounter{algorithm}% New algorithm
     \hrule height.8pt depth0pt \kern2pt% \@fs@pre for \@fs@ruled
     \renewcommand{\caption}[2][\relax]{% Make a new \caption
       {\raggedright\textbf{Procedure:} ##2\par}%
       \ifx\relax##1\relax % #1 is \relax
         \addcontentsline{loa}{algorithm}{\protect\numberline{\thealgorithm}##2}%
       \else % #1 is not \relax
         \addcontentsline{loa}{algorithm}{\protect\numberline{\thealgorithm}##1}%
       \fi
       \kern2pt\hrule\kern2pt
     }
  }{% \end{breakablealgorithm}
     \kern2pt\hrule\relax% \@fs@post for \@fs@ruled
   \end{center}
  }
\makeatother
\usepackage{enumerate}
\usepackage{appendix}

\cvprfinalcopy % *** Unco	mment this line for the final submission

\newcommand{\tabincell}[2]{\begin{tabular}{@{}#1@{}}#2\end{tabular}}

\begin{document}

\begin{figure*}
  © 2021 IEEE. Personal use of this material is permitted. Permission from IEEE must be obtained for all other uses, in any current or future media, including reprinting republishing this material for advertising or promotional purposes, creating new collective works, for resale or redistribution to servers or lists, or reuse of any copyrighted component of this work in other works.
\end{figure*}

\clearpage

%%%%%%%%% TITLE
\title{Novel Adaptive Binary Search Strategy-First Hybrid Pyramid- and Clustering-Based CNN Filter Pruning Method without Parameters Setting}

\author{Kuo-Liang Chung \qquad Yu-Lun Chang \qquad Bo-Wei Tsai  \\
\large National Taiwan University of Science and Technology \vspace{-.2em}\\
\large Department of Computer Science \& Information Engineering \vspace{-.2em}\\
\normalsize
\{klchung01,~lance030201,~haha4nima\}@gmail.com
}

\maketitle
%\thispagestyle{empty}
%%%%%%%%% ABSTRACT
\begin{abstract}
\vspace{-.5em}
Pruning redundant filters in CNN models has received growing attention. In this paper, we propose an adaptive binary search-first hybrid pyramid-  and clustering-based (ABSHPC-based) method for pruning filters automatically. In our method, for each convolutional layer, initially a hybrid pyramid data structure is constructed to store the hierarchical information of each filter. Given a tolerant accuracy loss, without parameters setting, we begin from the last convolutional layer to the first layer; for each considered layer with less or equal pruning rate relative to its previous layer, our ABSHPC-based process is applied to optimally partition all filters to clusters, where each cluster is thus represented by the filter with the median root mean of the hybrid pyramid, leading to maximal removal of redundant filters. Based on the practical dataset and the CNN models, with higher accuracy,  the thorough experimental results demonstrated the significant parameters and floating-point operations reduction merits of the proposed filter pruning method relative to the state-of-the-art methods.
\end{abstract}

% plain text
% Deeper neural networks are more difficult to train. We present a residual learning framework to ease the training of networks that are substantially deeper than those used previously. We explicitly reformulate the layers as learning residual functions with reference to the layer inputs, instead of learning unreferenced functions. We provide comprehensive empirical evidence showing that these residual networks are easier to optimize, and can gain accuracy from considerably increased depth. On the ImageNet dataset we evaluate residual nets with a depth of up to 152 layers---8x deeper than VGG nets but still having lower complexity. An ensemble of these residual nets achieves 3.57% error on the ImageNet test set. This result won the 1st place on the ILSVRC 2015 classification task. We also present analysis on CIFAR-10 with 100 and 1000 layers.

%The depth of representations is of central importance for many visual recognition tasks. Solely due to our extremely deep representations, we obtain a 28% relative improvement on the COCO object detection dataset. Deep residual nets are foundations of our submissions to ILSVRC & COCO 2015 competitions, where we also won the 1st places on the tasks of ImageNet detection, ImageNet localization, COCO detection, and COCO segmentation.

%%%%%%%%% BODY TEXT

\vspace{-1em}
\section{INTRODUCTION}
  Convolutional neural networks (CNN) have been widely used in developing deep learning models for many applications in computer vision, image processing, compression, speech processing, medical diagnosis, and so on. LeCun \textit{et al.} \cite{LeCun-1998} proposed the LeNet-5 model, which consists of three convolutional layers and two fully connected layers, for document recognition. Krizhevsky \textit{et al.} \cite{Krizhevsky-2012} proposed the AlexNet model consisting of five convolutional layers and three fully-connected layers to solve the visual object recognition problem in the ImageNet challenge \cite{Deng-2009}. Their AlexNet model needs a few million parameters (also called weights).\par

  Due to the great technology achievement in graphics processor units (GPU) with efficient parallel, pipeline, and vectorization processing capabilities, several interesting CNN models have been developed, such as VGG-16 \cite{Simonyan-2014}, SegNet \cite{Badrinarayanan-2017}, AlexNet \cite{Krizhevsky-2012}, GoogLeNet \cite{Szegedy-2014}, GAN (Generative Adversarial Network) \cite{Goodfellow-2014}, Mask-RCNN \cite{He-2017-RCNN}, U-Net \cite{Ronneberger-2015}, and so on. Among these developed CNN models, some may need more than several giga of parameters. However, some of these parameters are redundant, which leads to the model compression study. The compressed CNN models can thus be deployed into resource constrained embedding systems, such as mobile phones and surveillance systems \cite{Chen-2016}.\par

  In the past years, many model compression methods have been developed, including: (1) the weight pruning approach, (2) the layer pruning approach, (3) the  knowledge distillation approach, (4) the low-rank matrix factorization approach, and (5) the filter pruning approach. Two commonly used metrics to evaluate the model compression performance are the reduction rate of the number of parameters required in the compressed CNN model over the number of parameters required in the original CNN model, simply called the parameters reduction rate, and the reduction rate of the number of floating-point operations (FLOPs) used over the number of FLOPs used in the original CNN model, simply called the FLOPs reduction rate.\par

  In the weight pruning approach \cite{H. Cai-2018}, \cite{S. Han-2016}, \cite{S. Han-2015}, \cite{Y. He-2018}, \cite{LeCun-1990}, \cite{Simonyan-2014}, \cite{K. Wang-2019}, when one absolute weight value of the kernel in the filter is less than the specified threshold, it could be zeroized. However, due to the irregular weight zeroization for each filter, it may need a sparse matrix computation-supporting library to accelerate the related convolutional operations. Alternatively, we can quantize each weight value  by limited precision, where a lookup table shared by all the filters is often used to map the quantized weight value to an optimized integer. In the layer pruning approach \cite{Li-2017}, \cite{Chen-2016}, \cite{Chen-2019}, researchers suggested pruning all the filters in the considered convolutional layer.\par

  Chen and Zhao \cite{Chen-2019} analyzed the feature representations in different layers, and then a feature diagnosis approach was proposed to prune unimportant layers. Finally, the compressed model was retrained by the knowledge distillation technique \cite{Hinton-2015} to compensate for the performance loss. Cheng \textit{et al.} \cite{Cheng-2017} pointed out that purely using the knowledge distilling approach in model compression is not suitable for solving the classification-oriented problems.\par

  The low-rank factorization technique \cite{Denil-2013} was proposed to decompose the weight matrix as a product of two smaller matrices by controlling the rank of the weight matrix such that many parameter values can be predicted, and then those redundant parameter values can be pruned. Because more and more 3x3 and 1x1 kernels have been used in the current models \cite{Chandra-2016}, \cite{Yang-2019}, it limits the parameters and FLOPs reduction rates by the low-rank factorization technique.\par

\begin{figure}[]
  \centering
  \includegraphics[height=5.5cm]{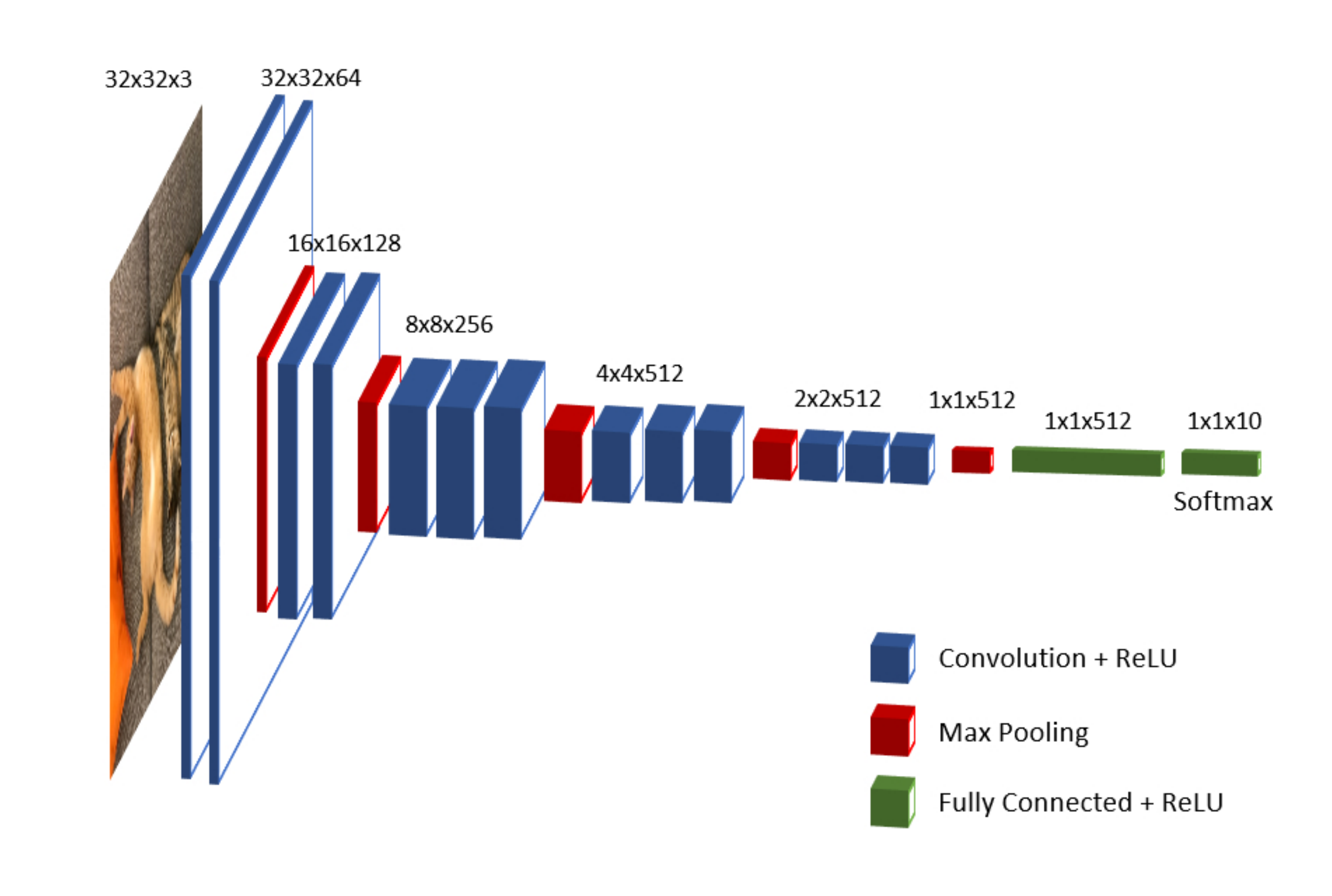}
  \caption{The VGG-16 model.}
  \label{fig:VGG}
\end{figure}

  Due to user accessibility and friendly tuning, the filter pruning approach provides the efficiency benefit on both CPU and GPU because no special hardware and/or library supports are required. In the next subsection, several state-of-the-art filter pruning methods are introduced. For easy exposition, we take VGG-16 (Visual Geometry Group-16) \cite{VGG16-A}, as shown in Fig. \ref{fig:VGG}, as the example in the introduction of the related work. In VGG-16, there are thirteen convolutional layers, namely Conv1-Conv13, and one fully connected layer, namely Fc1. The configuration of VGG-16 is shown in Table \ref{table:layer_config}.

\subsection{Related Work}
  In Li \textit{et al.}'s method  \cite{Li-2017}, for each convolution layer, they sorted all filters according to their absolute weight sums in increasing order. Next, based on a fixed filter pruning rate, namely 50\%, for the 8th-13th layers, i.e. Conv8-Conv13, and the first layer Conv1, they discarded those filters with smaller absolute sums. However, due to the fixed pruning rate setting, it limits the filter pruning performance. Based on the filter sparsity concept in \cite{C.T. Liu-2018}, Liu \textit{et al.} \cite{Liu-2019} defined a filter as being more redundant than others when that filter has several coefficients which are less than the mean value of all absolute filter weights in that layer. By using the rate-distortion optimization technique in image coding, they proposed a computation-performance optimization approach to prune redundant filters. Due to the available code, Li \textit{et al.}'s fixed pruning rate- and backward pruning-based (FPBP-based) method \cite{Li-2017}, simply called the FPBP method, is included in the comparative methods.\par

  Given an allowable number of filters to be  pruned for each layer, He \textit{et al.} \cite{He-2017} considered the distortion between the original feature map and the resultant feature map caused by the pruning filters, and then they derived a 1-norm regularization formula to model the filter pruning problem as a constraint minimization problem. Experimental results demonstrated the accuracy merit of their method relative to other methods \cite{Jaderberg-2014}, \cite{Zhang-2014}. Lin \textit{et al.} \cite{Lin-2020} modeled the filter pruning problem as a minimization problem associated with an objective function problem to seek the best tradeoff between the filter selection and the minimization of the cross-entropy loss for classification error between  the labels of ground truth and the output of the last layer in the considered CNN model. Luo \textit{et al.} \cite{Luo-2017}, \cite{Luo-2019} first calculated the sum of all entries of each channel in the feature map produced by the $i$th convolutional layer, and then they pruned the channel with the minimal sum. The pruning process is repeated until the specified channel pruning rate is reached; the subsequent removal of the corresponding filters in the $i$th and ($i$+1)th layers is followed.\par

  In He \textit{et al.}’s soft filter pruning (SFP) method \cite{He-2018}, for the $i$th layer, they first sorted all filters in the layer according to their $L_{2}$-norm values. Next, according to a fixed pruning rate, namely 25\% for the 1st-13th layers, i.e. Conv1-Conv13, they zeroized these filters with smaller $L_{2}$-norm values. In the next retraining step, all the determined filters including the zeroized filters are retrained. Following the same pruning rate, the above step is repeated until the number of epochs has been reached. Finally, they discarded those filters still with zero $L_{2}$-norm values. Their SFP method can not only be applied to maintain the model capacity to achieve better model compression performance, but it is also less dependent on the pre-trained model. However, the same fixed pruning rate setting for each convolutional layer limits the filter pruning performance. Due to the available code, the SFP method is included in the comparative methods.\par

  In \cite{Ayinde-2018}, based on the cosine-based similarity metric to measure the similarity level between two filter clusters in the $i$th layer, if the similarity value is larger than the specified distance threshold, namely 0.3 empirically, the two clusters are merged. Ayinde and Zurada \cite{Ayinde-2018} repeat their cosine-based merging method (CMM) until all similar clusters are merged. For each cluster, they randomly select one filter to represent that cluster and discard the remaining filters in that cluster. In addition, they remove the corresponding feature maps produced by those discarded filters in the $i$th layer; in the ($i$+1)th layer, they also discard the filters corresponding to the removed feature maps produced by the $i$th layer. However, the fixed distance threshold setting for determining the clusters for each convolutional layer limits the pruning performance. Due to the available code, the CMM method \cite{Ayinde-2018} is included in the comparative methods.\par

  To improve the previous SFP method, He \textit{et al.} \cite{He-2019} proposed a geometric median-based filter pruning (GMFP) method. For the considered layer with $k$ filters, they first calculate the geometric center of all the filters in that layer, where the sum of all distances between each filter and the geometric center is the smallest among that for the other location. Then, according to a specified pruning rate, namely 30\% for layers 1-13, they zeroize the $k$*30\% filters which are closest to the geometric center. In the subsequent retraining step, all the filters are retained. The above GMFP process and the retraining step are repeated until the required number of epochs has been reached. The experimental results  justified better parameters and FLOPs reduction merits by the GMFP method relative to the SFP method \cite{He-2018}. However, the fixed pruning rate setting for discarding redundant filters for each convolutional layer limits the pruning performance. Due to the available codes, the GMFP method is included in the comparative methods.
  \subsection{Motivation} \label{sec:IC}
    The above-mentioned limitation existing in the related filter pruning work prompted us to develop an automatically adaptive filter pruning method to achieve significant reduction of parameters and FLOPs required in the CNN models relative to the related state-of-the-art methods.\par

  \subsection{Contributions} \label{sec:IC}
    In this paper, without parameters setting, we propose an automatically adaptive binary search-first hybrid pyramid- and clustering-based (ABSHPC-based) filter pruning method to effectively remove redundant filters for CNNs, achieving significant parameters reduction and FLOPs reduction effects. The four contributions of this paper are clarified as follows.\par

    In the first contribution, given an allowable accuracy loss, namely 0.5\%, based on the CIFAR-10, we take VGG-16 as the representative CNN model. From the constructed accuracy-pruning rate curves shown in Fig. \ref{fig:color_accuracy}, three observations are delivered, and these observations prompted us to prune filters following the order from the last convolutional layer to the first layer. According to this backward pruning order, without parameters setting, we propose an automatically adaptive filter pruning method such that the pruning rates comply with a decreasing sequence.\par

    In the second contribution, we propose a novel hybrid pyramid (HP) data structure to store the hierarchical information of each filter in the considered convolutional layer, where the root mean of HP indicates the absolute sum of the absolute weights of that filter, and then all HPs in the considered layer are sorted in increasing order based on their root means. Futhormore, for the considered layer, under the same accuracy loss, we propose an ABSHPC-based filter pruning process to remove the redundant filters to achieve the maximal pruning rate. Empirically, our method begins with the 13th convolutional layer Conv13, and the maximal pruning rate of this layer is $\alpha_{13}$ = 87.5\%.\par

    In the third contribution, for the next convolutional layer, namely Conv12, the initial pruning rate of Conv12, namely $\alpha_{12}$, is equal to $\alpha_{13}$. Based on this initial pruning rate $\alpha_{12}$ and the same allowable accuracy loss of 0.5\%, the proposed ABSHPC-based filter pruning process is applied to discard the redundant filters in Conv12 as much as possible. Empirically, it yields $\alpha_{12}$ = 87.5\%. We repeat the above  ABSHPC-based filter pruning processes for Conv11, Conv10, Conv9, ..., Conv2, and Conv1, where the resultant eleven filter pruning rates are 87.5\%, 87.5\%, 62.5\%, 62.5\%, 50.5625\%, 31.25\%, 31.25\%, 0\%, 0\%, 0\%, and 0\%, respectively. \par

    In the fourth contribution, with the highest accuracy, the parameters reduction rate gains of our filter pruning method over the four state-of-the-art methods, namely the FPBP method \cite{Li-2017}, the SFP method \cite{He-2018}, the CMM method \cite{Ayinde-2018}, and the GMFP method \cite{He-2019}, are 24.35\%, 44.55\%, 24.67\%, and 37.94\%, respectively; the FLOPs reduction rate gains of our method over the four methods are 17.78\%, 8.33\%, 7.93\%, and 1.46\%, respectively. In addition, based on the same dataset on AlexNet \cite{Krizhevsky-2012}, our method also achieves substantial parameters and FLOPs reduction merits when compared with the related methods. \par

    The rest of this paper is organized as follows. In Section II, three observations on the accuracy-pruning rate curves for all convolutional layers are delivered. In Section III, the HP data structure is proposed to store the hierarchical information of each filter. Then, a fast HP-based closest filter finding operation is proposed. In Section IV, the proposed ABSHPC-based filter pruning process for each convolutional layer is presented. Then, the whole procedure of our filter pruning method is described. In Section V, the thorough experimental results are illustrated to justify the parameters and FLOPs reduction merits of our filter pruning method. In Section VI, some concluding remarks are addressed.

\section{THREE OBSERVATIONS ON THE CONSTRUCTED ACCURACY-PRUNING RATE CURVES}

   \label{sec:II}
  Based on the CIFAR-10 dataset, in which 50000 32x32 images are used as the training set and the disjoint 10000 32x32 images are used as the testing set, and VGG-16, based on our experiments with 20 epochs, three observations on the constructed accuracy-pruning rate curves are presented.\par

\begin{figure}[]
  \centering
  \includegraphics[height=6.3cm]{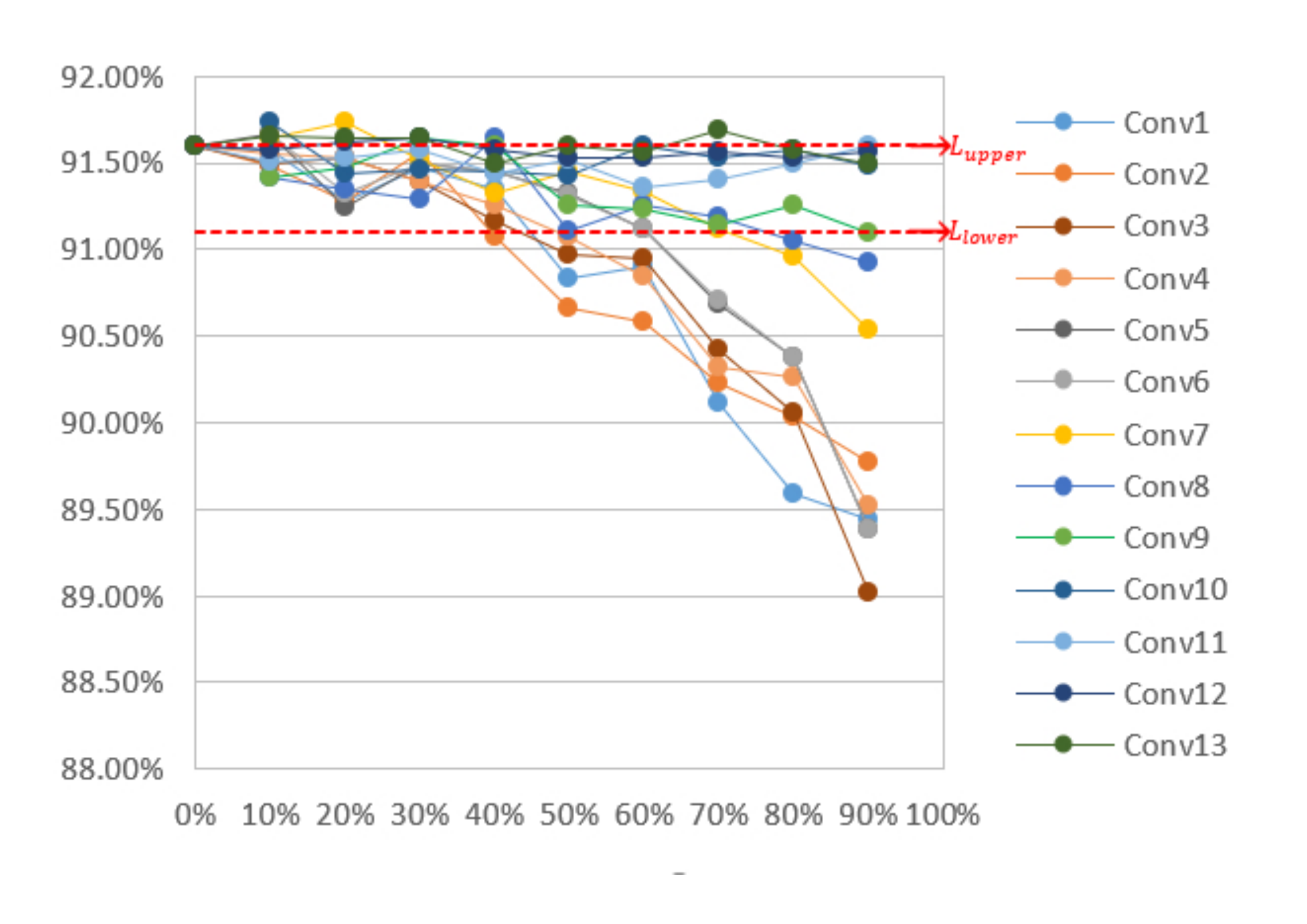}
  \caption{The accuracy-pruning rate curves of the thirteen convolutional layers for VGG-16 and CIFAR-10.}
  \label{fig:color_accuracy}
\end{figure}

  According to the ten pruning rates \cite{Li-2017}, namely 0\%, 10\%, 20\%, 30\%, ..., and 90\%, for each convolutional layer while retaining all the filters for the other twelve layers each time. Based on the above pruning rates setting, the filters with low absolute sums are pruned first. As a result, the thirteen accuracy-pruning rate curves are depicted in Fig. \ref{fig:color_accuracy} in which the X-axis denotes the pruning rate and the Y-axis denotes the accuracy value. Note that without pruning any filters for each convolutional layer, the classification accuracy of the trained VGG-16 model is 91.60\%, as depicted by the dashed line $L_{upper}$ of Fig. \ref{fig:color_accuracy}, indicating the accuracy upper bound. \par

  Suppose the accuracy loss is 0.5\%. As depicted in Fig. \ref{fig:color_accuracy}, the dashed line $L_{lower}$ denotes the accuracy lower bound 91.10\% (= 91.60\% - 0.5\%). From Fig. \ref{fig:color_accuracy} and the visual help of $L_{lower}$ and $L_{upper}$, three new observations are given; they are: (1) the seven convolutional layers, Conv7, Conv8, Conv9, Conv10, Conv11, Conv12, and Conv13, form a group and each of them can tolerate higher pruning rates rather than the other layers, even more than the filter pruning rate 50\% used in \cite{Li-2017}, (2) instead of setting a zero pruning rate for Conv6-Conv2 \cite{Li-2017}, nonzero pruning rates can be considered for these layers, (3) for Conv13-Conv1, their filter pruning rates could form a decreasing sequence.

\begin{table}[]
\centering
\caption{\protect\centering \protect\linebreak THE CONFIGURATION OF THE THIRTEEN CONVOLUTIONAL LAYERS IN VGG-16.}
\vspace{6pt}
\begin{tabular}{|c|c|c|}
  \hline
  \rowcolor[HTML]{C0C0C0}
  Layer          & Filter (\#Filters)          & Feature Map           \\ \hline
  Conv1          & 3x3x3 (64)               & 32x32x64              \\ \hline
  Conv2          & 3x3x64 (64)              & 32x32x64              \\ \hline
  Maxpool        & -                    & 16x16x64              \\ \hline
  Conv3          & 3x3x64 (128)              & 16x16x128             \\ \hline
  Conv4          & 3x3x128 (128)             & 16x16x128             \\ \hline
  Maxpool        & -                    & 8x8x128               \\ \hline
  Conv5          & 3x3x128 (256)             & 8x8x256               \\ \hline
  Conv6          & 3x3x256 (256)             & 8x8x256               \\ \hline
  Conv7          & 3x3x256 (256)             & 8x8x256               \\ \hline
  Maxpool        & -                    & 4x4x256               \\ \hline
  Conv8          & 3x3x256 (512)              & 4x4x512               \\ \hline
  Conv9          & 3x3x512 (512)             & 4x4x512               \\ \hline
  Conv10         & 3x3x512 (512)             & 4x4x512               \\ \hline
  Maxpool        & -                    & 2x2x512               \\ \hline
  Conv11         & 3x3x512 (512)             & 2x2x512               \\ \hline
  Conv12         & 3x3x512 (512)             & 2x2x512               \\ \hline
  Conv13         & 3x3x512 (512)             & 2x2x512               \\ \hline
  Maxpool        & -                    & 1x1x512               \\ \hline
  Fc1            & -                    & 1x1x512               \\ \hline
  Fc2        & -                    & 1x1x10                \\ \hline
  \end{tabular}
\label{table:layer_config}
\end{table}

\par

\section{HYBRID PYRAMID-BASED FILTER REPRESENTATION AND THE CLOSEST FILTER FINDING OPERATION}

   \label{sec:III}
  In this section, we first propose a HP data structure to store the hierarchical information of each filter in the convolutional layer. Next, based on the proposed HP data structure, some inequalities are derived to explain why given a filter as a key, its closest filter in a considered filter set can be found quickly. Note that the closest filter finding operation plays an important role in the proposed HP-based clustering process, which will be presented in Section IV.A.

\subsection{Hybrid Pyramid-Based Filter Representation} \label{sec:IIIA}
  \textit{1) Constructing HP for each filter:} We first take the 13th convolutional layer, namely Conv13, as the example to explain how to construct the HP data structure to represent the hierarchical information of each filter in Conv13. Our proposed HP is different from the Laplacian pyramid and the quadtree pyramid \cite{Burt-1983}, \cite{LeCun-1990}, \cite{Lin-2001} used in coding.\par

\begin{figure}[]
  \centering
    \subfigure{\includegraphics[height = 6cm]{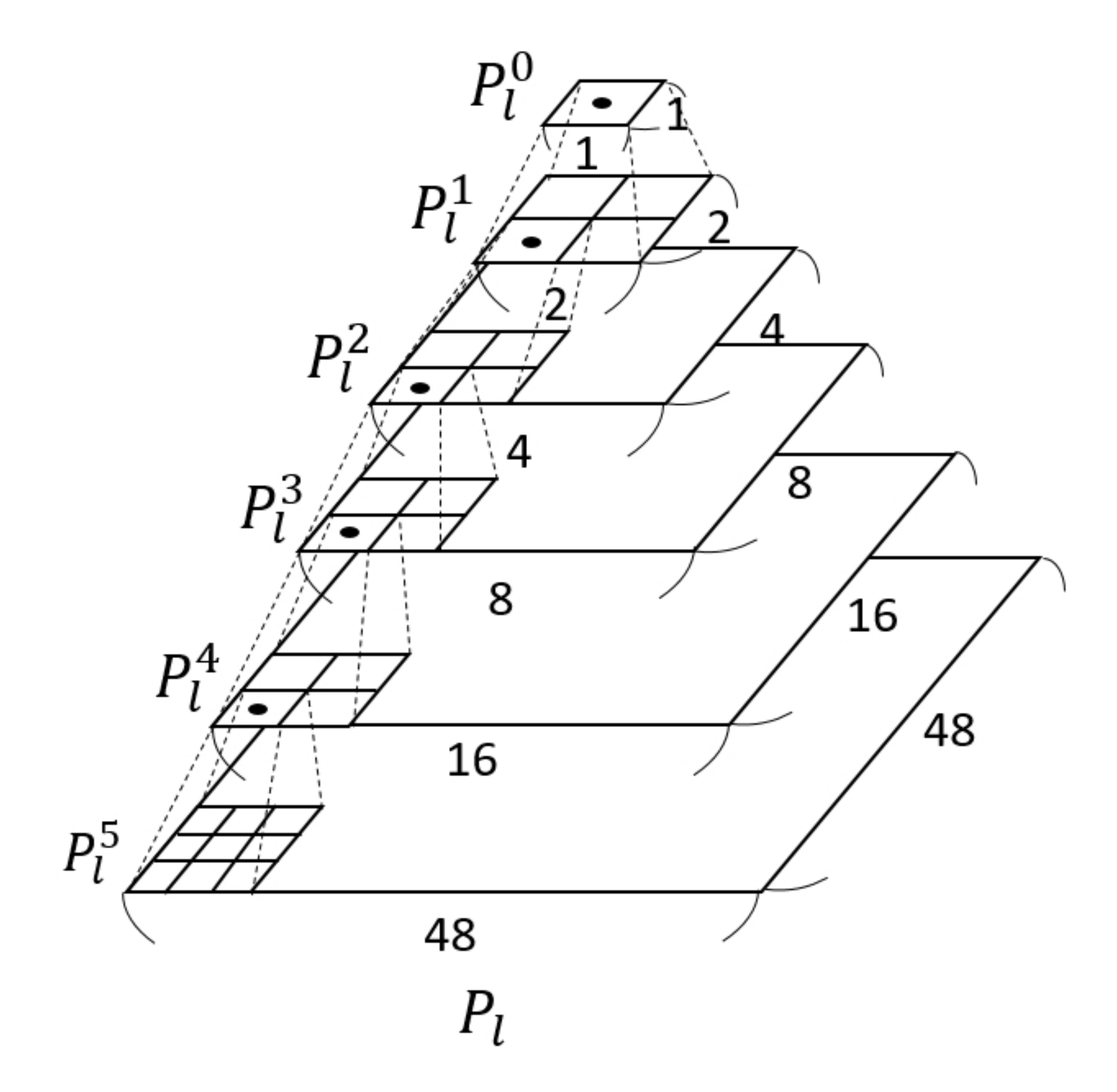}}
    \subfigure{\includegraphics[height = 4.4cm]{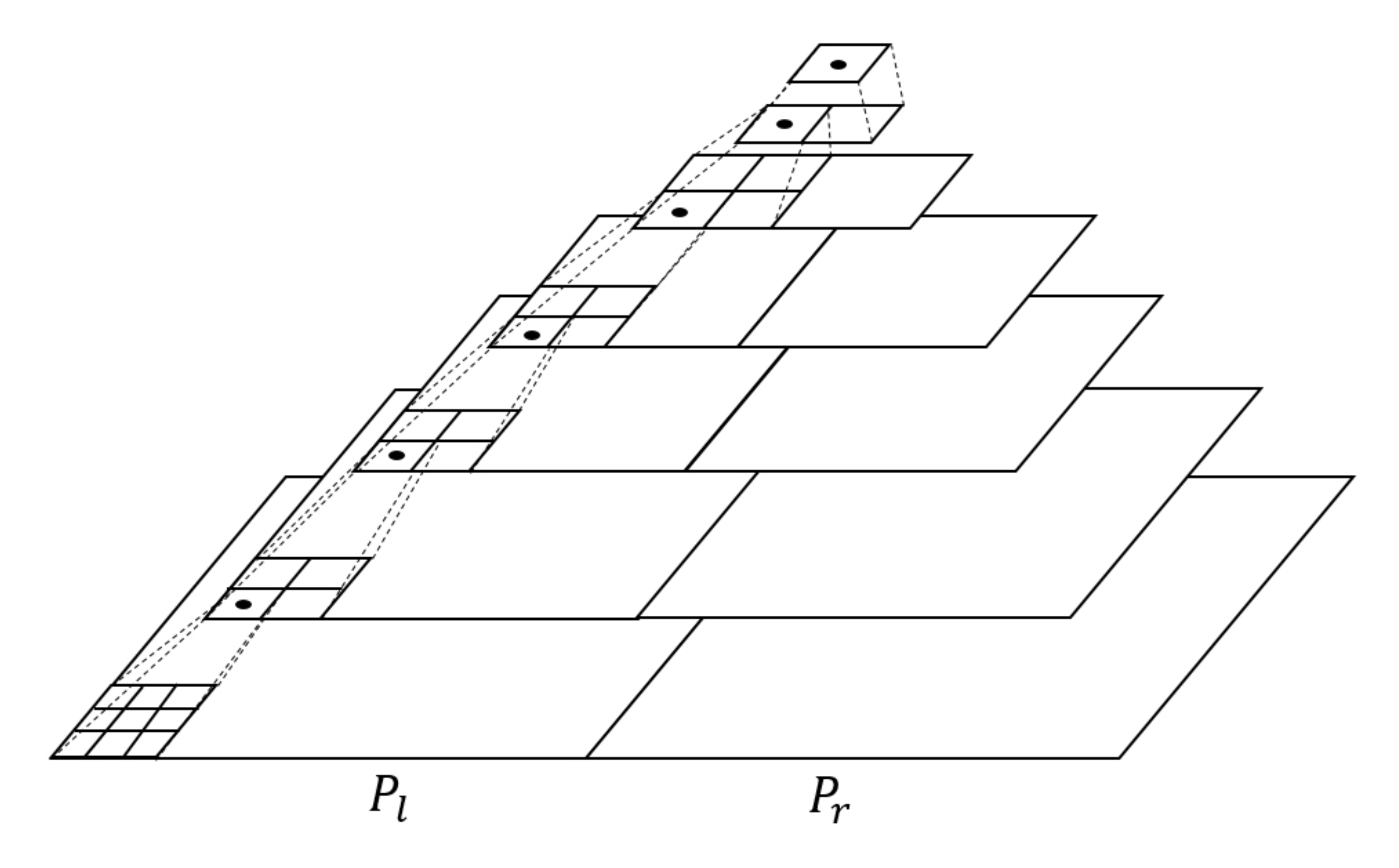}}
    \caption{
    The constructed hybrid pyramid for each 3x3x512 filter. (a) The constructed left sub-pyramid $P_l$ for the 48x48 matrix converted from the former 256 3x3 kernels in the filter. (b) The constructed hybrid pyramid $P$ by connecting the two sub-pyramids $P_{l}$ and $P_{r}$.
    }
  \label{fig:pyramid}
  \end{figure}

  In Table \ref{table:layer_config}, Conv13 consists of 512 3x3x512 filters, where each contains 512 channels in which each channel is exactly a 3x3 kernel. Initially, we take absolute operation on each weight in the filter to make the weight value nonnegative. For each filter, the 512 3x3 kernels are denoted by $K_1$, $K_2$, ..., and $K_{512}$. Among the 512 kernels, the former 256 kernels, $K_1$, $K_2$, ..., and $K_{256}$, form a square 48x48 matrix, denoted by $M_l$, in which the first kernel $K_1$ is located at the top-left corner of $M_l$ and the kernel $K_{256}$ is located at the bottom-right corner.  In the same way, the latter 256 kernels,
  $K_{257}$, $K_{258}$, ..., and $K_{512}$, form a square 48x48 matrix, where the kernels $K_{257}$ and $K_{512}$ are located at the top-left and bottom-right corners of  $M_r$, respectively. The 48x48 matrix $M_l$ constitutes the base of the left sub-pyramid $P_l$, as shown in Fig. \ref{fig:pyramid}(a); $M_r$ constitutes the base of the right sub-pyramid $P_r$. Connecting the two sub-pyramids, $P_l$ and $P_r$, the constructed HP for representing each 3x3x512 filter is depicted in Fig. \ref{fig:pyramid}(b).

  As depicted in Fig. \ref{fig:pyramid}(a), the left sub-pyramid $P_{l}$ consists of six levels, $P^{0}_{l}$, $P^{1}_{l}$, ..., and $P^{5}_{l}$, where the fifth level $P^{5}_{l}$ denotes the 48x48 matrix $M_l$, forming the base of $P_{l}$; after averaging each 3x3 sub-matrix of $P^{5}_{l}$ to a mean value, the 4th level $P^{4}_{l}$ is constructed to store the condensed 16x16 matrix; the root level $P_l^0$ saves the absolute mean value of $P^{5}_l$.
  In the same way, the right sub-pyramid $P_{r}$ is constructed to store the hierarchical information of the considered 3x3x256 filter. Finally, the roots of $P_l$ and $P_r$, i.e. $P^{0}_{l}$ and $P^{0}_{r}$, are connected by a common root to construct a hybrid pyramid. Fig. \ref{fig:pyramid}(b) depicts the resultant HP for saving the hierarchical information of each 3x3x512 filter in Conv13.\par

  \textit{2) Computational complexity analysis and the sorted HPs for all filters in the layer:} We first analyze the computational complexity for constructing the hybrid pyramid of each filter in Conv13, and then analyze the computational complexity for constructing the sorted HPs.\par

  For the fifth level of $P_l$, namely $P^5_l$, its size is $N^{2}$ = $48^{2}$. In terms of the big-O complexity notation \cite{Cormen-2009}, it is not hard to verify that it takes $O$($N^2$) (= 4/3*$N^{2}$ + constant) time to construct the sub-pyramid $P_l$. Similarly, it takes $O(N^2)$ time to construct $P_r$. Consequently, it takes $O$($N^2$) time to construct the HP, as depicted in Fig. \ref{fig:pyramid}(b), for saving the hierarchical information of each 3x3x512 filter in Conv13.\par

\begin{figure}[]
\centering
  \includegraphics[height = 6.0cm]{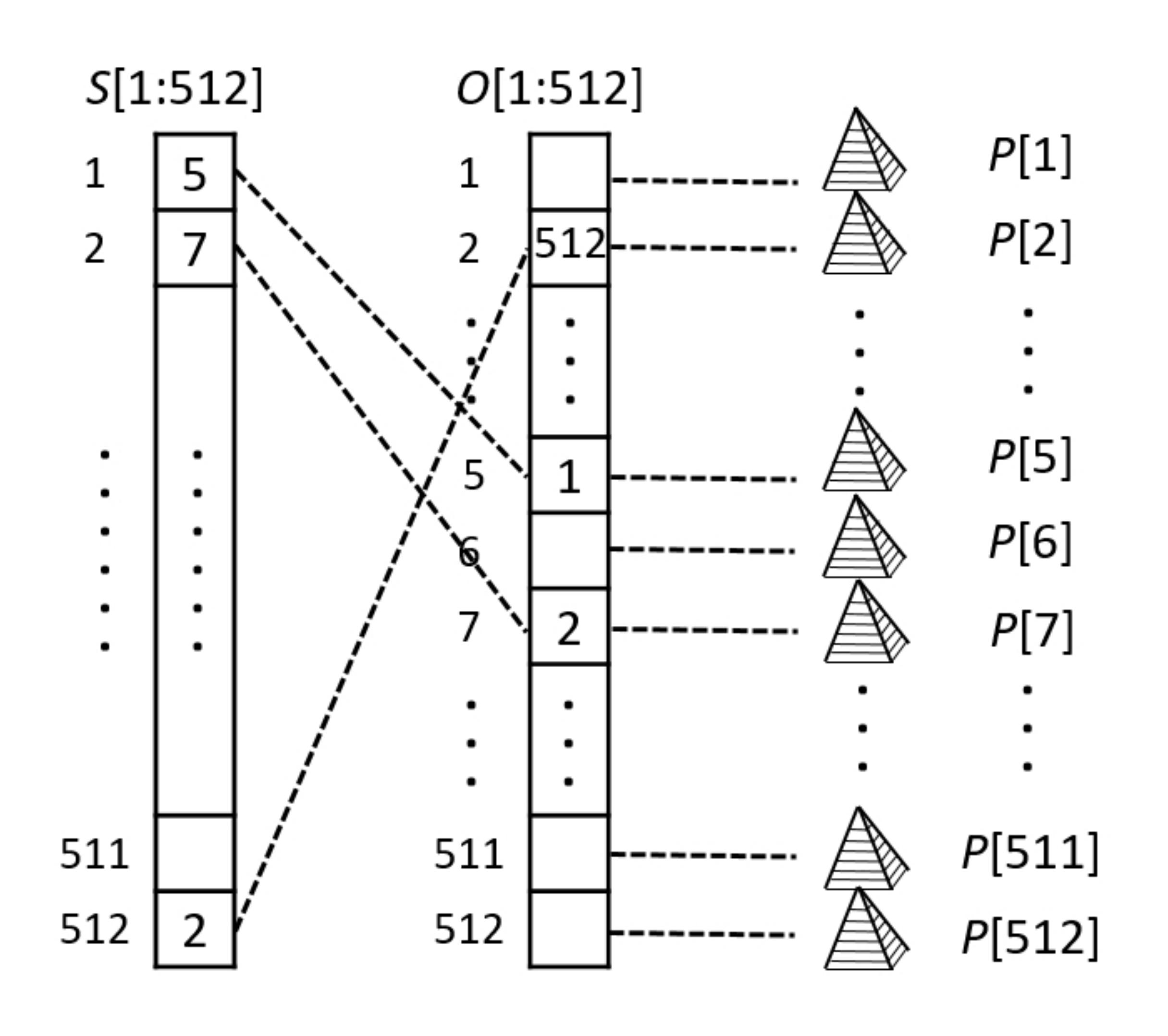}
  \caption{
  The constructed hybrid pyramids for the 512 3x3x512 filters in the 13th convolutional layer of VGG-16.
  }
\label{fig:sorted_order}
\end{figure}

  According to the above HP construction method for each filter, the constructed 512 HPs for the 512 filters in Conv13 are depicted in Fig. \ref{fig:sorted_order}, where the 512 HPs are denoted by $P[1]$, P$[2]$, ..., and $P[512]$ corresponding to the filters $F[1]$, $F[2]$, ..., and $F[512]$, respectively.\par

  According to the 512 root means of $P[1]$, $P[2]$, ..., and $P[512]$, the 512 HPs are sorted in increasing order; $S[i]$, 1 $\leq$ $i$ $\leq$ 512, saves the original index of the sorted HP which is in the $i$th place.
  This above sorting job can be done in $O$($|\textit{\textbf{F}}|$ $\log$ $|\textit{\textbf{F}}|$) time, where \textit{\textbf{F}} =\{$F[1]$, $F[2]$, ..., $F[512]$\} and $|\textit{\textbf{F}}|$ (= 512). As shown in Fig. \ref{fig:sorted_order}, “$S[1]$ = 5” indicates that the index of the HP with the smallest root mean is 5 corresponding to $P[5]$; “$S[2]$ = 7” and “$S[512]$ = 2'' indicate that the indices of the HPs with the second smallest and the largest root means are 7 and 2  corresponding to $P[7]$ and $P[2]$, respectively. Considering the inverse of $S[i]$ (= $j$), we build up the array $O[j]$ (= $i$) to access the sorted order of the hybrid pyramid $P[j]$, 1 $\leq$ $j$ $\leq$ 512. Following the above three examples, we have $O[2]$ = 512, $O[5]$ =1, and $O[7]$ = 2.

According to Table \ref{table:layer_config} for VGG-16, the number of the constructed HPs for all the filters in each convolutional layer and the number of levels required for each HP are tabulated in Table \ref{table:pyramid_nums}, in which ``\#(Hybrid Pyramids)'' denotes the number of HPs required in each convolutional layer and ``\#(Levels)'' denotes the number of levels required for each constructed HP.

\begin{table}[]
    \centering
      \caption{\protect\centering \protect\linebreak THE NUMBER OF HYBRID PYRAMIDS AND LEVELS FOR EACH CONVOLUTIONAL LAYER IN VGG-16.}
      \vspace{6pt}
      \begin{tabular}{|c|c|c|}
        \hline
        \rowcolor[HTML]{C0C0C0}
        Layer No.     & \#(Hybrid Pyramids)     & \#(Levels)            \\ \hline
        1             & 64                      & 2                      \\ \hline
        2             & 64                      & 5                     \\ \hline
        3             & 128                     & 5                     \\ \hline
        4             & 128                     & 6                     \\ \hline
        5             & 256                     & 6                     \\ \hline
        6-7           & 256                     & 6                     \\ \hline
        8             & 512                     & 6                     \\ \hline
        9-13          & 512                     & 7                     \\ \hline
        \end{tabular}
\label{table:pyramid_nums}
\end{table}

\subsection{Fast Hybrid Pyramid-Based Closest Filter Finding} \label{sec:IIIB}
  In this subsection, based on the sorted HPs for the considered convolutional layer, given a filter $F[i]$ as a key, some inequalities are first derived to assist in quickly finding its closest filter in the considered filter set.

  \textit{1) Proof of inequalities and its application to prune unnecessary $L_{2}$-norm distance calculations between two filters:} Given a 3x3x512 filter $F[i]$ as a key corresponding to the hybrid pyramid $P[i]$, let the root mean of $P[i]$ be denoted by $P^{0}[i]$ which is equal to the mean of the two root means $P^{0}_{l}[i]$ and $P^{0}_{r}[i]$. Let the $L_{2}$-norm squared distance between $P^{0}[i]$ and $P^{0}[j]$ be denoted by $d^{2}(P^{0}[i], P^{0}[j])$ where $P^{0}[j]$ denotes the root mean of a possible closest 3x3x512 filter candidate $F[j]$ in the considered filter set with respect to $F[i]$. We have the following inequality.

\begin{lemma}
  \label{label:label1}
  2*$d^{2}(P^{0}[i], P^{0}[j])$ $\leq$ $d^{2}((P^{0}_{l}[i]$, $P^{0}_{r}[i])$,  ($P^{0}_{l}[j]$, $P^{0}_{r}[j]$).
\end{lemma}
  \proof See Appendix I.\\[0.05cm]

  The physical meaning behind Lemma \ref{label:label1} can be highlighted by an example. For example, suppose the $L_2$-norm squared distance $d^{2}$($P^{0}[i]$, $P^{0}[j]$) is equal to 4, and then the value of $2*d^{2}$($P^{0}[i]$, $P^{0}[j]$) is equal to 8. By Lemma \ref{label:label1}, theoretically, the value of $d^{2}$(($P^{0}_{l}[i]$, $P^{0}_{r}[i]$), ($P^{0}_{l}[j]$, $P^{0}_{r}[j]$)) must be larger than or equal to 8, even though we do not calculate its true $L_{2}$-norm squared distance value. \par

  We now extend Lemma \ref{label:label1} to derive the inequalities for the same level between two hybrid pyramids to prune the unnecessary $L_2$-norm squared distance calculation between $F[i]$ and $F[j]$ in a top-down manner, achieving fast closest filter finding of $F[i]$ in the considered filter set.\par

  From Table \ref{table:pyramid_nums}, each of the five convolutional layers, Conv9-Conv13, has the same number of HPs, namely 512, and each HP has seven levels. Similar to the proving technique for Lemma \ref{label:label1}, we have a more general result.

  \begin{theorem}
  \label{theorem:theorem1}
  For the 9th-13th convolutional layers of VGG-16, we have the following inequalities:
  \begin{equation}\label{eq:VGG_layer9-13}
  \begin{aligned}
    2*4^{4}*9*{d}^{2}({P}^{0}[i], {P}^{0}[j]) \leq  \\
    4^{4}*9*d^{2}((P^{0}_{l}[i], P^{0}_{r}[i]), (P^{0}_{l}[j], P^{0}_{r}[j])) \leq \\
    4^{3}*9*d^{2}((P^{1}_{l}[i], P^{1}_{r}[i]), (P^{1}_{l}[j], P^{1}_{r}[j])) \leq  ... \leq \\
    d^{2}((P^{5}_{l}[i],P^{5}_{r}[i]), (P^{5}_{l}[j],P^{5}_{r}[j]))
  \end{aligned}
  \end{equation}
  \proof See Appendix II.\\[0.05cm]
  \end{theorem}

\noindent{We also explain the physical meaning behind Theorem \ref{theorem:theorem1} by one example. Let the considered filter set be denoted by $\textit{\textbf{F}}$$'$ and let the temporary closest filter of $F[i]$ be $F[k]$ $\in$ $\textit{\textbf{F}}$$'$.
Let the $L_{2}$-norm squared distance between $F[i]$ and $F[k]$ be 16384. We now examine whether the other filter $F[j]$ $\in$ $\textit{\textbf{F}}$$'$ can replace $F[k]$ as a better closest filter candidate of $F[i]$. Assume the $L_{2}$-norm square distance between the root mean of $P[i]$ and the root mean of $P[j]$ is 4, i.e. $d^{2}$($P^{0}[i]$, $P^{0}[j]$) = 4, and then we immediately know $2*4^4*9*d^{2}$($P^{0}[i]$, $P^{0}[j]$) = $2*4^4*9*4$ = 18432. Because of $d^{2}$($F[i]$, $F[k]$) = 16384 < $2*4^4*9*d^2$($P^{0}[i]$, $P^{0}[j]$) = 18432, by Theorem \ref{theorem:theorem1}, we know that theoretically, the $L_{2}$-norm squared distance $d^{2}$(($P^{5}_{l}[i]$, $P^{5}_{r}[i]$), ($P^{5}_{l}[j]$, $P^{5}_{r}[j]$)) is always larger than or equal to $2*4^4*9*d^{2}$($P^{0}[i]$, $P^{0}[j]$) = 18432, so we ignore the true squared distance calculation for $d^{2}$(($P^{5}_{l}[i]$, $P^{5}_{r}[i]$), ($P^{5}_{l}[j]$, $P^{5}_{r}[j]$)) because the filter $F[j]$ has no chance of being a better closest filter of $F[i]$ relative to $F[k]$, leading to the computation reduction effect.}\par

After discussing how to apply Theorem \ref{theorem:theorem1} to reduce the computational complexity of the $L_2$-norm squared distance calculation between two filters in the closest filter finding for the $k$th, 9 $\leq$ $k$ $\leq$ 13, convolutional layer, we now derive the  inequalities for the $k$th, 6 $\leq$ $k$ $\leq$ 8, convolutional layer; as listed in Table \ref{table:pyramid_nums}, the constructed hybrid pyramid for each 3x3x256 filter is the same as in Fig. \ref{fig:pyramid}(a). In the same way, for any two filters $F[i]$ and $F[j]$ in the $k$th layer, it yields

\begin{equation}
\begin{aligned}
4^{4}*9*d^{2}(P^{0}[i],P^{0}[j]) \leq 4^{3}*9*d^{2}(P^{1}[i],P^{1}[j]) \\
\leq ... \leq d^{2}(P^{5}[i],P^{5}[j])
\label{eq:VGG_layer6-8}
\end{aligned}
\end{equation}
\\

Similarly, in the $k$th, 4 $\leq$ $k$ $\leq$ 5, convolutional layer, for any two 3x3x128 filters, $F[i]$ and $F[j]$, we have

\begin{equation}
\begin{aligned}
2*4^{3}*9*d^{2}(P^{0}[i],P^{0}[j]) \leq 4^{3}*9*d^{2}(P^{1}[i],P^{1}[j]) \\
\leq ... \leq d^{2}(P^{4}[i],P^{4}[j])
\label{eq:VGG_layer4-5}
\end{aligned}
\end{equation}
\\

For any two 3x3x64 filters, $F[i]$ and $F[j]$, in the $k$th, 2 $\leq$ $k$ $\leq$ 3,  convolutional layer, we have

\begin{equation}
\begin{aligned}
4^{3}*9*d^{2}(P^{0}[i],P^{0}[j]) \leq 4^{2}*9*d^{2}(P^{1}[i],P^{1}[j]) \\
\leq ... \leq d^{2}(P^{4}[i],P^{4}[j])
\label{eq:VGG_layer2-3}
\end{aligned}
\end{equation}
Finally, for any two 3x3x3 filters, $F[i]$ and $F[j]$, in the first convolutional layer, we have

\begin{equation}
\begin{aligned}
27*d^{2}(P^{0}[i],P^{0}[j]) \leq d^{2}(P^{1}[i],P^{1}[j])
\label{eq:VGG_layer1}
\end{aligned}
\end{equation}

In terms of equation number, Table \ref{table:layer_eq} tabulates the general inequalities for each convolutional layer in VGG-16, and these inequalities can be used to prune unnecessary calculations in the proposed HP-based closest filter finding operation.\par

\begin{table}[]
    \centering
      \caption{\protect\centering \protect\linebreak THE DERIVED INEQUALITIES FOR EACH CONVOLUTIONAL LAYER IN VGG-16.}
      \vspace{6pt}
\begin{tabular}{|c|c|c|}
  \hline
  \rowcolor[HTML]{C0C0C0}
  Layer No.     & Inequalities          \\ \hline
  1             & Eq. (\ref{eq:VGG_layer1})                              \\ \hline
  2-3           & Eq. (\ref{eq:VGG_layer2-3})                        \\ \hline
  4-5           & Eq. (\ref{eq:VGG_layer4-5})                  \\ \hline
  6-8           & Eq. (\ref{eq:VGG_layer6-8})                            \\ \hline
  9-13          & Eq. (\ref{eq:VGG_layer9-13})                            \\ \hline
  \end{tabular}
\label{table:layer_eq}
\end{table}

\textit{2) The proposed hybrid pyramid-based closest filter finding operation:} We still take Conv13 as the layer example. Given a filter $F[i]$ in that layer as the key and under the considered filter set $\textit{\textbf{F}}$$'$, the proposed fast closest filter finding operation wants to find the closest filter $F[j]$ in $\textit{\textbf{F}}$$'$ such that the $L_2$-norm squared distance between $F[i]$ and $F[j]$ is the smallest.

In the first step, all the HPs of the filters in $\textit{\textbf{F}}$$'$ are sorted in increasing order based on their root means, and the sorted HPs are corresponding to these filters $F'[S[1]]$, $F'[S[2]]$, ..., and $F'[S[|\textit{\textbf{F'}}|]]$. Given the root mean of the HP of $F[i]$ as a key, according to the binary search process, we can quickly find the closest root mean of the HP of $F'[S[k]]$ $\in$ $\textit{\textbf{F}}$$'$. Next, the squared distance between $F[i]$ and $F'[S[k]]$ is obtained by computing the $L_2$-norm squared distance between the base of the HP of $F[i]$ and the base of the HP of  $F'[S[k]]$ as the temporary minimum distance, denoted by $d^{2}_{min}$.\par

In the second step, for any other filter candidate $F'[S[m]]$, $m$ $\neq$ $k$, in $\textit{\textbf{F}}$$'$, corresponding to the hybrid pyramid $P'[S[m]]$, we further want to find the closest filter of $F[i]$ in a smaller filter set instead of examining all filters in $\textit{\textbf{F}}$$'$ - \{$F'[S[k]]$\}. In what follows, we explain how to modify Theorem \ref{theorem:theorem1} to derive a smaller search range for further reducing the number of filter candidates to be examined. When compared with the temporary minimum distance $d_{min}$, by Theorem \ref{theorem:theorem1}, the closest filter candidate of $F[i]$, namely $F'[S[m]]$, must satisfy the following inequality:

\begin{equation}
\begin{aligned}
  d^2_{min} \geq 2*4^4*9*d^2(P^0[i], P'^0[S(m)])  \\
   = 2*4^4*9*(P^{0}[i] - P'^{0}[S(m)])^2
\label{eq:bandwidth_1}
\end{aligned}
\end{equation}
\\

where $P'^{0}[S(m)]$ denotes the root mean of the hybrid pyramid of $F'[S[m]]$; $P^0[i]$ denotes the root mean of the hybrid pyramid of $F[i]$.\par

We first divide both sides of Eq. (\ref{eq:bandwidth_1}) by $2*4^4*9$, and then we take the square root operation on both sides. Considering the two possible cases, ($P^0[i]$- $P'^0[S(m)]$ $\geq$ 0) or ($P^0[i]$ - $P'^0[S(m)]$ $\leq$ 0), the smaller search range of the promising closest hybrid pyramids for $F[i]$ corresponding to the hybrid pyramid $P[i]$ is thus bounded by

\begin{equation}
\begin{aligned}
  (P^0[i] -   \frac{d_{min}}{\sqrt{2}*4^2*3}) \leq \\
    P'^0[S(m)]  \leq ( P^0[i] + \frac{d_{min}}{\sqrt{2}*4^2*3})
\label{eq:bandwidth_2}
\end{aligned}
\end{equation}
\\

The range in Eq. (\ref{eq:bandwidth_2}) is used to narrow the search range for finding the closest filter of $F[i]$. On the other hand, if the root mean of one filter $F'[S[m]]$ is out of the search range in Eq. (\ref{eq:bandwidth_2}), $F'[S[m]]$ will be viewed as a useless filter and will be kicked out immediately; otherwise, it goes downward to the next level of both hybrid pyramids $P[i]$ and $P'[S[m]]$ and checks whether the filter $F'[S[m]]$ should be rejected or should go downward to the next level. When it goes downward to the bottom level and the $L_2$-norm squared distance between the two related bases  is less than $d^2_{min}$, then the previous closest filter candidate $F’[S[k]]$ is replaced by the current filter $F'[S[m]]$ as the new closest filter candidate to $F[i]$. We repeat the above step until the true closest filter of $F[i]$ is found.

\section{THE PROPOSED AUTOMATICALLY ADAPTIVE BINARY SEARCH-FIRST HYBRID PYRAMID- AND CLUSTERING-BASED FILTER PRUNING METHOD} \label{sec:IV}
We first present the proposed HP-based clustering process, in which our HP-based closest filter finding operation is used as a subroutine. Secondly, without parameters setting, we present the whole procedure of the proposed automatical ABSHPC-based filter pruning method.\par

\subsection{The Proposed Hybrid Pyramid-Based Clustering Process} \label{sec:IVA}

In the considered convolutional layer, let the currently considered filter set be denoted by $\bar{F}$. Suppose the filter pruning rate of this layer is $\frac{c}{|\bar{F}|}$.  On the other hand, the goal of the proposed HP-based clustering process is to partition all the filters in $\bar{F}$ into c clusters such that one suitable filter in each cluster is selected as the representative of that cluster, achieving the filter pruning effect.\par

First, we randomly select $c$ filters from $\bar{F}$ as the initial $c$ clusters, denoted by $\bar{F}^c$, where each cluster contains only one selected filter. We take each filter $F[i]$ in $\bar{F}$ - $\bar{F}^c$ as a key, and then using the proposed HP-based closest filter finding operation, which has been described in Subsection III.B.2, we can quickly find the closest filter of $F[i]$, namely $F[j]$ in $\bar{F}^c$.\par

Next, we group those filters belonging to the same cluster as a new cluster, and then for each new cluster, the filter with the median root mean of the hybrid pyramid is selected as the representative filter of that cluster. Therefore, each cluster is represented by such a representative filter, and we discard the other filters in that cluster. On the other hand, each cluster now contains only one representative filter. In our experience, instead of taking the mean filter of all filters in that cluster as the representative, the above median root mean-oriented selection strategy has better filter pruning performance due to the selection of the highly distinctive representative. After reconstructing the $c$ clusters via these c representative filters, we repeat the above clustering process to refine the $c$ clusters until there is no change to the representative of each cluster. Finally, in these convergent $c$ clusters, for each cluster, we take the filter with the median root mean as the representative of that cluster, and prune the other filters in that cluster.\par

\subsection{The Whole Procedure of the Proposed Automatical ABSHPC-Based  Filter Pruning Method} \label{sec:IVB}
After presenting our HP-based clustering process, we now present the proposed automatical ABSHPC-based filter pruning method for the thirteen convolutional layers in VGG-16 and the whole procedure is shown below.\par
\vspace{3pt}

\begin{breakablealgorithm}
  \renewcommand{\thealgorithm}{}
\caption{Automatical ABSHPC-Based Filter Pruning}
\begin{raggedright}
\KwIn{Training set CIFAR-10, Trained VGG-16 with the accuracy 91.60\%, and the allowable accuracy loss 0.5\%.}
\KwOut{Compressed VGG-16.}
\end{raggedright}

\label{protocol1}
%\noindent\rule{\linewidth}{0.7pt}
\begin{enumerate}[{\textbf{Step \arabic{enumi}}}.]

\item (\textbf{initialization for binary search}) Perform $k$ $:=$ 13, $R^{(13)}_{upper}$ $:=$ 1, $R^{(13)}_{lower}$ $:=$ 0, $R^{(13)}$ $:=$ 0, $R^{(1)}_{upper}$ $:=$ 1, $R^{(1)}_{lower}$ $:=$ 0, and $R^{(1)}$ $:=$ 0.

\item (\textbf{construct the sorted hybrid pyramids for the \textit{\textbf{k}}th layer}) Construct the HP for each filter in the \textit{k}th convolutional layer. Next, sort all these HPs in increasing order based on their root means. Let the initial set of all filters in the \textit{k}th layer be denoted by $F^{(k)}$ and let $N^{(k)}$(= $|F^{(k)}|$) denote the number of all filters in the $k$th layer.

\item (\textbf{For 12 $\geq$ \textbf{\textit{k}} $\geq$ 1, based on the pruning rate passed by the last layer, perform the HP-based filter pruning process once}) If $k$ = 13, go to Step 4; otherwise, based on the pruning rate $R^{(k)}$ := $\frac{|F^{(k+1)}|}{N^{(k+1)}}$ obtained in the last convolutional layer, we apply the proposed HP-based clustering process to partition the current filter set $F^{(k)}$ into $c$ (= $R^{(k)}|F^{(k)}|$) clusters. For each cluster, we select its representative filter with the median root mean and discard the other filters in that cluster. Let all the representatives of the $c$ clusters be denoted by $F^{(k)}$.
After retraining VGG-16 based on the current filter set $F^{(k)}$ in the $k$th layer and the stationary filters in the other layers, if the accuracy loss is larger than 0.5\%, we set $R^{(k)}_{upper}$ := 1 and $R^{(k)}_{lower}$ := $\frac{|F^{(k+1)}|}{N^{(k+1)}}$, conceptually moving the current binary search cursor to the right to increase the number of representative filters, and go to Step 4; otherwise, go to Step 5.

\item (\textbf{Adaptive binary search-first HP-based filter pruning}) Let $R^{(k)}_{old}$ := $R^{(k)}$ and $R^{(k)}$ := $\frac{R^{(k)}_{upper}+R^{(k)}_{lower}}{2}$. If $|R^{(k)}_{old}$ - $R^{(k)}|$ is less than 0.0125, it means that the binary search process has been done for six rounds, and then we go to Step. 5; otherwise, we apply the HP-based clustering process to partition the current filter set $F^{(k)}$ into $c$ (=$R^{(k)}|F^{(k)}|)$ clusters.
For each cluster, we select its representative filter and discard the other filters in that cluster. Let the set of these representatives of the $c$ clusters still be denoted by $F^{(k)}$. After retraining VGG-16 based on $F^{(k)}$ and the stationary filters in the other layers, if the accuracy loss is larger than 0.5\%, we perform $R^{(k)}_{lower}$ := $\frac{R^{(k)}_{upper}+R^{(k)}_{lower}}{2}$ to move the current binary search cursor to the right to increase the number of representative filters in the next round, and then we go to Step 4; otherwise, we perform $R^{(k)}_{upper}$:= $\frac{R^{(k)}_{upper}+R^{(k)}_{lower}}{2}$ to move the current binary search cursor to the left to decrease the number of the representative filters in the next round and go to Step 4.

\item (\textbf{termination test}) If $k$ = 1, we report the compressed VGG-16 as the output and stop the procedure; otherwise, we perform $k$ := $k-1$ and go to Step 2.

\end{enumerate}
\end{breakablealgorithm}

\par

\begin{table*}[]
  \centering
    \caption{\protect\centering \protect\linebreak THE PARAMETERS AND FLOPS REDUCTION MERITS OF THE PROPOSED METHOD FOR VGG-16.}
    \vspace{6pt}
    \begin{tabular}{|c|c|c|c|c|c|c|}
      \hline
      \rowcolor[HTML]{C0C0C0}
      Method     & \#(Parameters) &\tabincell{c}{Parameters\\Reduction Rate}  & \#(FLOPs)   &\tabincell{c}{FLOPs\\Reduction Rate} &Accuracy &Accuracy loss      \\ \hline
      Baseline   & 14.90M          & 0\%     & 626.90M      &0\%  &91.60\%   &0\%                 \\ \hline
      %He \textit{et al.} \cite{He-2017} & -   & -   & -   & -   & -              \\ \hline
      FPBP \cite{Li-2017} &5.36M &64.00\% &412.5M &34.20\% &91.53\% &0.07\%         \\ \hline
      CMM \cite{Ayinde-2018} &5.41M &63.68\% &350.8M &44.05\% &91.56\% &0.04\%      \\ \hline
      SFP \cite{He-2018} &8.37M &43.80\% &353.3M &43.65\% &91.55\% &0.05\%         \\ \hline
      GMFP \cite{He-2019} &7.39M &50.41\% &310.2M &50.52\% &91.55\% &0.05\%         \\ \hline
      Ours &\textbf{1.74M} &\textbf{88.35\%} &\textbf{301M} &\textbf{51.98\%} &\textbf{91.57\%} &\textbf{0.03\%}                                      \\ \hline

    \end{tabular}
\label{table:result_vgg}
\end{table*}

\begin{table*}[]
  \centering
    \caption{\protect\centering \protect\linebreak THE FILTER PRUNING RATE OF EACH CONVOLUTIONAL LAYER FOR VGG-16.}
    \vspace{6pt}
    \resizebox{\textwidth}{!}{
    \begin{tabular}{|c|c|c|c|c|c|c|c|c|c|c|c|c|c|}
      \hline
      \rowcolor[HTML]{C0C0C0}
      Purning Rate (Layer)  &13 &12 &11 &10 &9 &8 &7 &6 &5 &4 &3 &2 &1      \\ \hline
      FPBP \cite{Li-2017} &50\% &50\% &50\% &50\% &50\% &50\% &0\% &0\% &0\% &0\% &0\% &0\% &50\%      \\ \hline
      CMM \cite{Ayinde-2018} &60.55\% &60.35\% &67.18\% &62.11\% &27.73\% &16.99\% &4.69\% &3.91\% &13.67\% &11.72\% &43.75\% &53.13\% &0\%      \\ \hline
      SFP \cite{He-2018} &25\% &25\% &25\% &25\% &25\% &25\% &25\% &25\% &25\% &25\% &25\% &25\% &25\%      \\ \hline
      GMFP \cite{He-2019} &30\% &30\% &30\% &30\% &30\% &30\% &30\% &30\% &30\% &30\% &30\% &30\% &30\%      \\ \hline
      Ours &87.5\% &87.5\% &87.5\% &87.5\% &62.5\% &62.5\% &50.5625\% &31.25\% &31.25\% &0\% &0\% &0\% &0\%  \\ \hline
    \end{tabular}}
  \label{table:VGG_each_layer}
\end{table*}

\section{EXPERIMENTAL RESULTS} \label{sec:IV}
  Based on the CIFAR-10 dataset and the two CNN models, VGG-16 and AlexNet, the comprehensive experiments are carried out to show the parameters and FLOPs reduction merits of our automatical ABSHPC-based filter pruning method relative to the state-of-the-art methods. Under the Windows 10 platform, the source code of our filter pruning method is implemented by Python language and can be accessed from \cite{Code}.\par
  All experiments are implemented using a desktop with an Intel Core i7-7700 CPU running at 3.6 GHz with 24 GB RAM and a Nvidia 1080Ti GPU. The operating system is Microsoft Windows 10 64-bit. The program development environment is the Python programming language.\\[0.05cm]

  \subsection{The Parameters and FLOPs Reduction Merits for VGG-16}

  Table \ref{table:result_vgg} tabulates the parameters and FLOPs reduction rates comparison among our ABSHPC-based filter pruning method and the four state-of-the-art methods \cite{Ayinde-2018}, \cite{He-2018}, \cite{He-2019}, \cite{Li-2017}. In detail, Table \ref{table:VGG_each_layer} tabulates the filter pruning rate of each convolutional layer by each considered method.\par
  Table \ref{table:result_vgg} indicates that by the baseline method without pruning any filters, the number of required parameters, denoted by \#(Parameters), the number of required FLOPs, denoted by \#(FLOPs), and the accuracy are 14.90M, 626.90M, and 91.60\%, respectively. In Table \ref{table:result_vgg}, with the highest accuracy and the lowest accuracy loss, our filter pruning method has the highest parameters and FLOPs reduction rates in boldface relative to the four state-of-the-art methods. In detail, the parameters reduction rate gains of our method over FPBP \cite{Li-2017}, SPF \cite{He-2018}, CMM \cite{Ayinde-2018}, and GMFP \cite{He-2019} are 24.35\%, 44.55\%, 24.67\%, and 37.94\%, respectively; the FLOPs reduction rate gains of our method over the four state-of-the-art methods are 17.78\%, 8.33\%, 7.93\%, and 1.46\%, respectively.\\[0.05cm]

  \subsection{The Parameters and FLOPs Reduction Merits for AlexNet}\par

  We first outline the configuration of AlexNet. Next, for each convolutional layer, the number of HPs and the number of levels of each HP is analyzed. Furthermore, the inequalities for each convolutional layer are provided. Finally, the parameters and FLOPs reduction merits of our ABSHPC-Based filter pruning method are demonstrated.\par

  1) \textit{The configuration of AlexNet}: AlexNet consists of five convolutional layers and two fully connected layers. Table \ref{table:layer_config_alex} tabulates the configuration of AlexNet in which there are 96 filters, each filter with size $11\times11\times3$, in Conv1; there are 256 filters, each filter with size $5\times5\times96$, in Conv2; there are 384 filters, each filter with size $3\times3\times256$, in Conv3;  there are 384 filters, each filter with size $3\times3\times384$, in Conv4; there are 256 filters, each filter with size $3\times3\times384$, in Conv5. \par

  \begin{table}[]
      \centering
        \caption{\protect\centering \protect\linebreak THE CONFIGURATION OF THE THIRTEEN CONVOLUTIONAL LAYERS IN ALEXNET.}
        \vspace{6pt}
  \begin{tabular}{|c|c|c|}
    \hline
    \rowcolor[HTML]{C0C0C0}
    Layer          & Filter (\#Filters)          & Feature Map           \\ \hline
    Conv1          & 11x11x3 (96)               & 32x32x96              \\ \hline
    Conv2          & 5x5x96 (256)              & 8x8x256             \\ \hline
    Maxpool        & -                        & 4x4x256               \\ \hline
    Conv3          & 3x3x256 (384)             & 4x4x384               \\ \hline
    Conv4          & 3x3x384 (384)             & 4x4x384               \\ \hline
    Conv5          & 3x3x384 (256)             & 4x4x256               \\ \hline
    Maxpool        & -                    & 2x2x256               \\ \hline
    Fc1            & -                    & 1x1x4096               \\ \hline
    Fc2            & -                    & 1x1x4096               \\ \hline
    Fc3        & -                    & 1x1x10                \\ \hline
    \end{tabular}
  \label{table:layer_config_alex}
  \end{table}

  2) \textit{The number of hybrid pyramids and the number of levels of each HP in each convolutional layer}: According to the configuration of AlexNet, as shown in Table \ref{table:layer_config_alex}, the number of the constructed HPs for all the filters in each convolutional layer and the number of levels required for each HP are tabulated in Table \ref{table:pyramid_nums_alex}. \par

  For the first layer, it is known that the number of HPs required for the first layer is 96, and each filter is of size $11\times11$$\times3$; the HP data structure of each filter connects three sub-HPs in which the base of each is a $11\times11$ matrix. Therefore, the level of each HP is three. For the second layer, the number of HPs required for the second layer is 256, and each filter is of size $5\times5$$\times96$; the HP data structure of each filter connects six sub-HPs in which the base of each sub-HP is a ($2^2$$\times5$)$\times$($2^2\times5$) matrix. Therefore, the level of each HP is five. To reduce the paper length, we omit the related discussion for Conv3-Conv5.

  \begin{table}[]
      \centering
        \caption{\protect\centering \protect\linebreak THE NUMBER OF HYBRID PYRAMIDS AND LEVELS FOR EACH CONVOLUTIONAL LAYER IN ALEXNET.}
        \vspace{6pt}
        \begin{tabular}{|c|c|c|}
          \hline
          \rowcolor[HTML]{C0C0C0}
          Layer No.     & \#(Hybrid Pyramids)     & \#(Levels)            \\ \hline
          1             & 96                      & 3                      \\ \hline
          2             & 256                     & 5                     \\ \hline
          3-4             & 384                     & 6                     \\ \hline
          5             & 256                     & 6                     \\ \hline
          \end{tabular}
  \label{table:pyramid_nums_alex}
  \end{table}

  3) \textit{The inequalities for each convolutional layer}: In terms of equation number, Table \ref{table:layer_eq_alex} tabulates the derived inequalities for each convolutional layer in AlexNet, and these inequalities can be used to prune unnecessary calculations in the proposed ABSHPC-based filter pruning method. \par
  Considering the first layer, from the constructed HP of each filter and the number of levels of each HP, as shown in Table \ref{table:pyramid_nums_alex}, according to the similar proving technique used in Theorem \ref{theorem:theorem1}, we have the following inequalities:

  \begin{equation}
  \begin{aligned}
  3*11^{2}*d^{2}(P^{0}[i],P^{0}[j]) \leq \\ 11^{2}*d^{2}((P^{0}_{1}[i],P^{0}_{2}[i],P^{0}_{3}[i]),(P^{0}_{1}[j],P^{0}_{2}[j],P^{0}_{3}[j])) \leq \\
  d^{2}((P^{1}_{1}[i],P^{1}_{2}[i],P^{1}_{3}[i]),(P^{1}_{1}[j],P^{1}_{2}[j],P^{1}_{3}[j]))
  \label{eq:alex_layer1}
  \end{aligned}
  \end{equation}

  In the same way, for the second, third, fourth, and fifth layers, the corresponding inequalities are given in
  Eq. (\ref{eq:alex_layer2}), Eq. (\ref{eq:alex_layer3}), Eq. (\ref{eq:alex_layer4-5}), and Eq. (\ref{eq:alex_layer4-5}), respectively.

  \begin{equation}
  \begin{aligned}
  6*2^{2}*5*d^{2}(P^{0}[i], P^{0}[j]) \leq \\
  2^{2}*5*d^{2}((P^{0}_{1}[i], ... , P^{0}_{6}[i]), (P^{0}_{1}[j], ... , P^{0}_{6}[j])) \leq \\
  2*5*d^{2}((P^{1}_{1}[i], ... , P^{1}_{6}[i]),  (P^{1}_{1}[j], ... , P^{1}_{6}[j])) \leq ... \leq \\
  d^{2}((P^{4}_{1}[i], ... , P^{4}_{6}[i]), (P^{4}_{0}[j], ... , P^{4}_{6}[j]))
  \label{eq:alex_layer2}
  \end{aligned}
  \end{equation}

  \begin{equation}
  \begin{aligned}
  4^{4}*9*d^{2}(P^{0}[i],P^{0}[j]) \leq 4^{3}*9*d^{2}(P^{1}[i],P^{1}[j])\\
  \leq ... \leq d^{2}(P^{5}[i],P^{5}[j])
  \label{eq:alex_layer3}
  \end{aligned}
  \end{equation}

  \begin{equation}
  \begin{aligned}
  6*2^{3}*3*d^{2}(P^{0}[i], P^{0}[j]) \leq \\
  2^{3}*3*d^{2}((P^{0}_{1}[i], ... , P^{0}_{6}[i]), (P^{0}_{1}[j], ... , P^{0}_{6}[j])) \leq \\
  2^{2}*3*d^{2}((P^{1}_{1}[i], ... , P^{1}_{6}[i]), (P^{1}_{1}[j], ... , P^{1}_{6}[j])) \leq ... \leq \\
  d^{2}((P^{5}_{1}[i], ... , P^{5}_{6}[i]), (P^{5}_{0}[j], ... , P^{5}_{6}[j]))
  \label{eq:alex_layer4-5}
  \end{aligned}
  \end{equation}
\vspace{2pt}

  \begin{table}[]
      \centering
        \caption{\protect\centering \protect\linebreak THE DERIVED INEQUALITIES FOR EACH CONVOLUTIONAL LAYER IN ALEXNET.}
        \vspace{6pt}
  \begin{tabular}{|c|c|c|}
    \hline
    \rowcolor[HTML]{C0C0C0}
    Layer No.     & Inequalities          \\ \hline
    1             & Eq. (\ref{eq:alex_layer1})                              \\ \hline
    2             & Eq. (\ref{eq:alex_layer2})                        \\ \hline
    3             & Eq. (\ref{eq:alex_layer3})                  \\ \hline
    4-5           & Eq. (\ref{eq:alex_layer4-5})                            \\ \hline
    \end{tabular}
  \label{table:layer_eq_alex}
  \end{table}
  \begin{table*}[]
    \centering
      \caption{\protect\centering \protect\linebreak THE PARAMETERS AND FLOPS REDUCTION MERITS OF THE PROPOSED METHOD FOR ALEXNET.}
      \vspace{6pt}
      \begin{tabular}{|c|c|c|c|c|c|c|}
        \hline
        \rowcolor[HTML]{C0C0C0}
        Method     & \#(Parameters) &\tabincell{c}{Parameters\\Reduction Rate}  & \#(FLOPs)   &\tabincell{c}{FLOPs\\Reduction Rate} &Accuracy &Accuracy loss      \\ \hline
        Baseline   &24.78M         & 0\%     &291.13M     &0\%  &78.64\%   &0\%                 \\ \hline
        CMM \cite{Ayinde-2018} &23.32M      & 5.89\%     &181.61M     &37.62\%  &78.62\%   &0.02\%                 \\ \hline
        SFP \cite{He-2018} &21.92M         & 11.56\%     &188.04M     &35.41\%  &78.62\%   &0.02\%                 \\ \hline
        %He \textit{et al.} \cite{He-2019} & -         & -\%     & -     &-\%  &-\%   &-\%                 \\ \hline
        Ours &\textbf{19.21M}         & \textbf{22.49\%}     &\textbf{165.59M}     &\textbf{43.12\%}  &\textbf{78.64\%}   &\textbf{0\%}                 \\ \hline

      \end{tabular}
  \label{table:result_alex}
  \end{table*}

  \begin{table*}[]
    \centering
      \caption{\protect\centering \protect\linebreak THE FILTER PRUNING RATE OF EACH CONVOLUTIONAL LAYER FOR ALEXNET.}
      \vspace{6pt}%
      \begin{tabular}{|c|c|c|c|c|c|}
        \hline
        \rowcolor[HTML]{C0C0C0}
        Purning Rate (Layer)  &5 &4 &3 &2 &1      \\ \hline
        %Li \textit{et al.} \cite{Li-2017}  &-\% &-\% &-\% &-\% &-\%      \\ \hline
        CMM \cite{Ayinde-2018} &12.89\% &7.55\% &2.34\% &19.92\% &57.29\%      \\ \hline
        SFP \cite{He-2018} &27\% &27\% &27\% &27\% &27\%      \\ \hline
        %He \textit{et al.} \cite{He-2019} &-\% &-\% &-\% &-\% &-\%      \\ \hline
        Ours &78.13\% &34.18\% &34.18\% &29.91\% &24.3\%      \\ \hline
      \end{tabular}
    \label{table:Alex_each_layer}
  \end{table*}

  4) \textit{The parameters and FLOPs reduction merits}: Table \ref{table:result_alex} tabulates the parameters and FLOPs reduction rates comparison among our ABSHPC-based filter pruning method and the two comparative methods \cite{Ayinde-2018}, \cite{He-2018}. In detail, Table \ref{table:Alex_each_layer} tabulates the filter pruning rate of each convolutional layer by each considered method.\par

  Table \ref{table:result_alex} indicates that by the baseline method without pruning any filters, the values of \#(Parameters), \#(FLOPs), and the accuracy are 24.78M, 291.13M, and 78.64\%, respectively. In Table \ref{table:result_alex}, with the highest accuracy and the lowest accuracy loss, our filter pruning method has the highest parameters and FLOPs reduction rates in boldface relative to the CMM \cite{Ayinde-2018} and SFP \cite{He-2018}. In detail, the parameters reduction rate gains of our method over CMM and SFP are 16.6\% (= 22.49\% - 5.89\%) and 10.93\% (= 22.49\% - 11.56\%), respectively; the FLOPs reduction rate gains of our method over the two comparative methods are 5.5\% (= 43.12\% - 37.62\%) and 7.71\% (= 43.12\% - 35.41\%), respectively.

  \section{CONCLUSION} \label{sec:V}
     Without parameters setting, we have presented the proposed automatically adaptive binary search-first HP- and clustering-based (ABSHPC-based) filter pruning method. In the presentation, we first provide some observations on the constructed accuracy-pruning rate curves for convolutional layers, and then the observations prompt us to prune filters from the last convolutional layer with the highest pruning rate to the first layer with the lowest pruning rate. For each convolutional layer, we remove the redundant filters in each cluster by only retaining the selected filter with the median root mean of the HP. Based on the CIFAR-10 dataset and the VGG-16 and AlexNet models, the comprehensive experimental data demonstrated the substantial parameters and FLOPs reduction merits of the proposed ABSHPC-based filter pruning method relative to the state-of-the-art methods.\par

     Our future work is to apply our automatic ABSHPC-based filter pruning method on other backbones like ResNet \cite{He-2015}, DenseNet \cite{Huang-2018}, MobileNet \cite{Howard-2017}, and on larger datasets like ImageNet \cite{Deng-2009}. In addition, we want to compare the related experimental results with the newly published filter pruning methods, \cite{{Wang-2019}}, \cite{Chen-2020}. \\[0.25cm]

     \noindent
     APPENDIX I: THE PROOF OF LEMMA \ref{label:label1}.\par
     \vspace{3pt}
     Assume the above lemma is true. Equivalently, the above inequality can be written as

     \begin{equation}
       \begin{aligned}
     2*(\frac{P^{0}_l[i]+P^{0}_r[i]}{2}-\frac{P^{0}_l[j]+P^{0}_r[j]}{2})^{2} \leq \\
     (P^{0}_l[i]-P^{0}_l[j])^{2}+ (P^{0}_r[i]-P^{0}_r[j])^{2}
     \label{eq:eq8}
     \end{aligned}
   \end{equation}
     \\
     Eq. (\ref{eq:eq8}) can be rewritten as

     \begin{equation}
       \begin{aligned}
     2*(\frac{P^{0}_l[i]-P^{0}_l[j]}{2}+\frac{P^{0}_r[i]-P^{0}_r[j]}{2})^{2} \leq \\
     (P^{0}_l[i]-P^{0}_l[j])^{2}+ (P^{0}_r[i]-P^{0}_r[j])^{2}
     \label{eq:eq9}
     \end{aligned}
   \end{equation}
     \\
     Eq. (\ref{eq:eq9}) is further expressed as

     \begin{equation}
       \begin{aligned}
     ((P^{0}_l[i]-P^{0}_l[j])+(P^{0}_r[i]-P^{0}_r[j]))^{2} \leq \\
     2*((P^{0}_l[i]-P^{0}_l[j])^{2}+(P^{0}_r[i]-P^{0}_r[j]))^{2}
     \label{eq:eq10}
     \end{aligned}
     \end{equation}
     \\
     Finally, Eq. (\ref{eq:eq10}) is simplified as

     \begin{equation}
       \begin{aligned}
     0 \leq ((P^{0}_l[i]-P^{0}_l[j])-(P^{0}_r[i]-P^{0}_r[j]))^{2}
     \label{eq:eq11}
     \end{aligned}
     \end{equation}
     \\
     Eq. (\ref{eq:eq11}) is always true. We thus confirm that our original assumption is true, and we complete the proof.\\[0.25cm]

     \noindent
     APPENDIX II: THE PROOF OF THEOREM \ref{theorem:theorem1}.\par
     \vspace{3pt}
     We proceed to the deeper level and want to derive the inequality for the relation between $d^{2}$(($P^{0}_{l}[i]$, $P^{0}_{r}[i]$), ($P^{0}_{l}[j]$, $P^{0}_{r}[j]$)) and $d^{2}$(($P^{1}_{l}[i]$, $P^{1}_{r}[i]$), ($P^{1}_{l}[j]$, $P^{1}_{r}[j]$)). By the similar proving technique as in Lemma \ref{label:label1}, it yields

     \begin{equation}
      \begin{aligned}
     4*d^{2}((P^{0}_{l}[i], P^{0}_{r}[i]), (P^{0}_{l}[j], P^{0}_{r}[j])) \leq \\
     d^{2}((P^{1}_{l}[i], P^{1}_{r}[i]), (P^{1}_{l}[j], P^{1}_{r}[j]))
     \label{eq:eq12}
     \end{aligned}
     \end{equation}
     \\
     Combining Lemma \ref{label:label1} and Eq. (\ref{eq:eq12}), it yields

     \begin{equation}
       \begin{aligned}
     2*4*d^{2}(P^{0}[i], P^{0}[j]) \leq \\
     4*d^{2}((P^{0}_{l}[i], P^{0}_{r}[i]), (P^{0}_{l}[j], P^{0}_{r}[j])) \leq \\
     d^{2}((P^{1}_{l}[i], P^{1}_{r}[i]), (P^{1}_{l}[j], P^{1}_{r}[j]))
     \label{eq:eq13}
     \end{aligned}
     \end{equation}

     In Table \ref{table:layer_config}, for the 9th-13th convolutional layers of VGG16, the number of filters and each filter structure are the same. Therefore, given two 3x3x512 filters, $F[i]$ and $F[j]$, corresponding to the two hybrid pyramids, $P[i]$ and $P[j]$, respectively, Eq. (\ref{eq:eq13}) indicates that Theorem \ref{theorem:theorem1} holds. We complete the proof.\par

     \section{ACKNOWLEDGEMENT} \label{sec:VI}
       The authors appreciate the proofreading help of Ms. C. Harrington to improve the manuscript.

{
%\small
\footnotesize
%\fontsize{8.8pt}{9pt}\selectfont
\bibliographystyle{ieee}
\bibliography{YL_arxiv_release}

\begin{thebibliography}{10}\itemsep=-1pt

\bibitem{Ayinde-2018}
B. O. Ayinde and J. M. Zurada,
\newblock Building efficient convnets using redundant feature pruning.
\newblock {\em arXiv preprint arXiv:1802.07653}, 2018.

\bibitem{Badrinarayanan-2017}
V. Badrinarayanan, A. Kendall, and R. Cipolla.
\newblock SegNet: A deep convolutional encoder-decoder architecture for image segmentation.
\newblock {\em IEEE Transactions on Pattern Analysis and Machine Intelligence}, vol. 39, no. 12, pp. 2481-2495, Dec. 2017.

\bibitem{Burt-1983}
P. J. Burt and E. H. Adelson.
\newblock The Laplacian pyramid as a compact image code.
\newblock {\em IEEE Transactions on Communications}, vol. 31, no. 4, pp. 532-540, Apr. 1983.

\bibitem{H. Cai-2018}
H. Cai, L. Zhu, and S. Han.
\newblock ProxylessNAS: Direct neural architecture search on target task and hardware.
\newblock {\em arXiv preprint arXiv:1812.00332}, 2018.

\bibitem{Chandra-2016}
B. Chandra and R. K.  Sharma.
\newblock Fast learning in deep neural networks.
\newblock {\em Neurocomputing}, vol. 171, pp. 1205–1215, Jan. 2016.

\bibitem{Chen-2016}
C. F. Chen, G. G. Lee, V. Sritapan, and C. Y. Lin.
\newblock Deep convolutional neural network on iOS mobile devices.
\newblock {\em IEEE International Workshop on Signal Processing Systems}, pp. 130–135, Oct. 2016.

\bibitem{Chen-2019}
S. Chen and Q. Zhao.
\newblock Shallowing Deep Networks: Layer-wise Pruning based on Feature Representations.
\newblock {\em IEEE Transactions on Pattern Analysis and Machine Intelligence}, vol. 41, no. 12, pp. 3048-3056, Dec. 2019.

\bibitem{Cheng-2017}
Y. Cheng, D. Wang, P. Zhou, and T. Zhang.
\newblock A survey of model compression and acceleration for deep neural networks.
\newblock {\em arXiv preprint arXiv:1710.09282}, 2017.

\bibitem{Chen-2020}
Z. Chen, T. B. Xu, C. Du, C. L. Liu, and H.He.
\newblock Dynamical Channel Pruning by Conditional Accuracy Change for Deep Neural Networks.
\newblock {\em IEEE Transactions on Neural Networks and Learning Systems}, pp. 1-15, April. 2020.

\bibitem{Cormen-2009}
T. H. Cormen, C. E. Leiserson, R. L. Rivest, and C. Stein.
\newblock {\em Introduction to Algorithms (Asymptotic Notation)}, 3rd ed. London, U.K.: MIT Press, sec. 3.1, 2009.

\bibitem{Deng-2009}
J. Deng, D. Wei, R. Socher, L. J. Li, K. Li, and F. F. Li.
\newblock ImageNet: a large-scale hierarchical image database.
\newblock {\em IEEE Conference on Computer Vision and Pattern Recognition}, pp. 248–255, 2009.

\bibitem{Denil-2013}
M. Denil, B. Shakibi, L. Dinh, M. Ranzato, and N. de Freitas.
\newblock Predicting parameters in deep learning.
\newblock {\em International Conference on Neural Information Processing Systems}, pp. 2148–2156, 2013.

\bibitem{Code}
{\em Execution code.} Accessed: 26 Jan. 2019. [Online]. Available: ftp://140.118.175.164/Model\_Compression/Codes.

\bibitem{Goodfellow-2014}
I. Goodfellow, J. Pouget-Abadie, M. Mirza, B. Xu, D. Warde-Farley, S. Ozair, A. Courville, and Y. Bengio.
\newblock Generative adversarial nets.
\newblock {\em In Advances in Neural Information Processing Systems}, pp. 2672–2680, 2014.

\bibitem{S. Han-2016}
S. Han, H. Mao, and W. J. Dally.
\newblock Deep compression: Compressing deep neural networks with pruning, trained quantization and huffman coding.
\newblock {\em Proceedings of the International Conference on Learning Representations}, no. 6, pp. 1-14,2016.

\bibitem{S. Han-2015}
S. Han, J. Pool, J. Tran, and W. J. Dally.
\newblock Learning both weights and connections for efficient neural network.
\newblock {\em Advances in neural information processing systems}, pp. 1135-1143, 2015.

\bibitem{He-2017-RCNN}
K. He, G. Gkioxari, P. Dollar, and R. Girshick.
\newblock Mask r-cnn.
\newblock {\em IEEE International Conference on Computer Vision}, pp. 2980–2988, 2017.

\bibitem{He-2015}
K. He, X. Zhang, S. Ren, and J. Sun.
\newblock Deep residual learning for image recognition.
\newblock {\em arXiv preprint arXiv:1512.03385}, pp. 2015.

\bibitem{He-2018}
Y. He, G. Kang, X. Dong, Y. Fu, and Y. Yang.
\newblock Soft filter pruning for accelerating deep convolutional neural networks.
\newblock {\em International Joint Conferences on Artificial Intelligence}, pp. 2234–2240, 2018.

\bibitem{Y. He-2018}
Y. He, J. Lin, Z. Liu, H. Wang, L. J. Li, and S. Han.
\newblock Amc: Automl for model compression and acceleration on mobile devices.
\newblock {\em European Conference on Computer Vision}, pp. 784-800, 2018.

\bibitem{He-2019}
Y. He, P. Liu, Z. Wang, Z. Hu, and Y. Yang.
\newblock Filter pruning via geometric median for deep convolutional neural networks acceleration.
\newblock {\em IEEE Conference on Computer Vision and Pattern Recognition}, pp. 4340-4349, 2019.

\bibitem{He-2017}
Y. He, X. Zhang, and J. Sun.
\newblock Channel pruning for accelerating very deep neural networks.
\newblock {\em IEEE International Conference on Computer Vision}, pp. 1398-1402, 2017.

\bibitem{Hinton-2015}
G. E. Hinton, O. Vinyals, and J. Dean.
\newblock Distilling the knowledge in a neural network.
\newblock {\em NIPS Deep Learning and Representation Learning Workshop}, pp. 1-9, 2015.

\bibitem{Howard-2017}
A. G. Howard, M. zhu, B. Chen, and D. Kalenichenko.
\newblock MobileNets: Efficient Convolutional Neural Networks for Mobile Vision Applications.
\newblock {\em arXiv preprint arXiv:1704.04861}, 2018.

\bibitem{Huang-2018}
G. Huang, Z. Liu, L. Maaten, and K. Q. Weinberger.
\newblock Densely connected convolutional networks.
\newblock {\em arXiv preprint arXiv:1608.06993}, 2017.

\bibitem{Jaderberg-2014}
M. Jaderberg, A. Vedaldi, and A. Zisserman.
\newblock Speeding up convolutional neural networks with low rank expansions.
\newblock {\em arXiv preprint arXiv:1405.3866}, 2014.

\bibitem{Krizhevsky-2012}
M. Jaderberg, A. Vedaldi, and A. Zisserman.
\newblock ImageNet classification with deep convolutional neural networks.
\newblock {\em Conference on Neural Information Processing Systems}, pp. 1097–1105, 2012.

\bibitem{LeCun-1998}
Y. Lecun, L. Bottou, Y. Bengio, P. Haffner.
\newblock Optimal brain damage.
\newblock {\em Proceedings of the IEEE}, pp. 2278-2324, Vol.86, No.11, Nov. 1998.

\bibitem{LeCun-1990}
C. H. Lee and L. H. Chen.
\newblock A fast search algorithm for vector quantization using mean pyramid of codewords.
\newblock {\em IEEE Transactions on Comminucations}, vol. 43, no. 2/3/4, pp. 1697-1702, Feb. 1995.

\bibitem{Li-2017}
H. Li, A. Kadav, I. Durdanovic, H. Samet, and H. P. Graf.
\newblock Pruning filters for efficient convnets.
\newblock {\em International Conference on Learning Representations}, pp. 1-13, 2017.

\bibitem{Lin-2020}
S. Lin, R. Ji, Y. Li, C. Deng, and X. Li.
\newblock Toward compact convnets via structure-sparsity regularized filter prunin.
\newblock {\em IEEE Transactions on Neural Networks and Learning Systems}, vol. 31, no. 2, pp. 574-588, Feb. 2020.

\bibitem{Lin-2001}
S. J. Lin, K. L. Chung, and L. C. Chang.
\newblock An improved search algorithm for vector quantization using mean pyramid structure.
\newblock {\em Pattern Recognition Letters}, vol. 22, no. 3-4, pp. 373-379, Mar. 2001.

\bibitem{Liu-2019}
C. T. Liu, T. W. Lin, Y. H. Wu, Y. S. Lin, H. Lee, Y. Tsao, and S. Y. Chien.
\newblock Computation-performance optimization of convolutional neural networks with redundant filter removal.
\newblock {\em IEEE Transactions on Circuits and Systems I: Regular Papers}, vol. 66, no. 5, pp. 1908-1921, May 2019.

\bibitem{C.T. Liu-2018}
C. T. Liu, Y. H. Wu, Y. S. Lin, and S. Y. Chien.
\newblock Computation-performance optimization of convolutional neural networks with redundant kernel removal.
\newblock {\em International Symposium on Circuits and Systems}, pp. 1–5, May 2018.

\bibitem{Liu-2017}
Z. Liu, J. Li, Z. Shen, G. Huang, S. Yan, and C. Zhang.
\newblock Learning efficient convolutional networks through network slimming.
\newblock {\em IEEE International Conference on Computer Vision}, pp. 2755–2763, 2017.

\bibitem{Liu-2018}
Z. Liu, M. Sun, T. Zhou, G. Huang, and T. Darrell.
\newblock Rethinking the value of network pruning.
\newblock {\em arXiv preprint arXiv:1810.05270}, 2018.

\bibitem{Luo-2017}
J. H. Luo, J. Wu, and W. Lin.
\newblock Thinet: A filter level pruning method for deep neural network compression.
\newblock {\em arXiv preprint arXiv:1707.06342}, 2017.

\bibitem{Luo-2019}
J. H. Luo, H. Zhang, H. Y. Zhou, C. W. Xie, J. Wu, and W. Lin.
\newblock ThiNet: pruning CNN filters for a thinner net.
\newblock {\em IEEE Transactions on Pattern Analysis and Machine Intelligence}, vol. 41, no. 10, pp. 2525-2538, Oct. 2019.

\bibitem{Ronneberger-2015}
O. Ronneberger, P. Fischer, and T. Brox.
\newblock U-net: Convolutional networks for biomedical image segmentation.
\newblock {\em International Conference On Medical Image Computing \& Computer Assisted Intervention}, pp. 234–241, 2015.

\bibitem{Simonyan-2014}
K. Simonyan and A. Zisserman.
\newblock Very deep convolutional networks for large-scale image recognition.
\newblock {\em arXiv preprint arXiv:1409.1556}, 2014.

\bibitem{Szegedy-2014}
C. Szegedy, W. Liu, Y. Jia, P. Sermanet, S. Reed, D. Anguelov, D. Erhan, V. Vanhoucke, and A. Rabinovich.
\newblock Going deeper with convolutions.
\newblock {\em arXiv preprint arXiv:1409.4842}, 2014.

\bibitem{VGG16-A}
VGG-16 Architecture \url{https://www.cs.toronto.edu/~frossard/post/vgg16/}

\bibitem{K. Wang-2019}
K. Wang, Z. Liu, Y. Lin, J. Lin, and S. Han.
\newblock Haq: Hardware-aware automated quantization with mixed precision.
\newblock {\em IEEE Conference on Computer Vision and Pattern Recognition}, pp. 8612-8620, 2019.

\bibitem{Wang-2019}
W. Wang, C. Fu, J. Guo, D. Cai, and X. He.
\newblock COP: Customized Deep Model Compression via Regularized Correlation-Based Filter-Level Pruning.
\newblock {\em  International Joint Conference on Artificial Intelligence}, 2019.

\bibitem{Yang-2019}
W. Yang, L. Jin, S. Wang, Z. Cu, X. Chen, and L. Chen.
\newblock Thinning of convolutional neural network with mixed pruning.
\newblock {\em IET Image Processing}, vol. 13, no. 5, pp. 779-784, May 2019.

\bibitem{Zhang-2014}
X. Zhang, J. Zou, K. He, and J. Sun.
\newblock Accelerating very deep convolutional networks for classification and detection.
\newblock {\em IEEE Transactions on Pattern Analysis And Machine Intelligence}, vol. 38, no. 10, pp. 1943–1955, 2016.


\end{thebibliography}
}

\end{document}